\documentclass[letterpaper, 10 pt, conference]{ieeeconf}  

\IEEEoverridecommandlockouts                              

\usepackage{graphics} 
\usepackage{graphicx} 
\usepackage{amsmath} 
\usepackage{amssymb}
\usepackage{cite} 

\makeatletter
\let\NAT@parse\undefined
\makeatother

\usepackage[colorlinks,linkcolor=black,anchorcolor=black,citecolor=blue,urlcolor=blue,hyperfootnotes=true]{hyperref}   
\usepackage[all]{hypcap}

\title{\LARGE \bf
Hybrid Impedance--Admittance Control with Multi-Link Aerial Robot for Contact-Rich Surface Sliding Task
}

\author{Zicheng Luo\textsuperscript{*}, Maolin Lei\textsuperscript{*}, Jinjie Li, Yicheng Chen, Zicen Xiong and Moju Zhao
\thanks{\textsuperscript{*}Corresponding authors. All authors are with the Department of Mechanical-Engineering, The University of Tokyo, Bunkyo-ku, Tokyo 113-8656, Japan.
{(email:\tt\small \{zicheng, maolin-lei, jinjie-li, yicheng-chen, zicen-xiong, chou\}@dragon.t.u-tokyo.ac.jp})}
}

\begin{document}

\maketitle
\thispagestyle{empty}
\pagestyle{empty}

\begin{abstract}
Multi-link aerial robots can actively deform their articulated structures during flight, giving them strong potential for aerial manipulation. However, they still face substantial challenges in contact-rich aerial manipulation tasks such as surface sliding, which requires both disturbance robustness and compliance to uncertain surface geometry. Force-control strategies such as impedance and admittance control are commonly employed to address these requirements. Although impedance control can provide disturbance-resistant interaction and admittance control can offer compliant adaptation, their opposite force--motion causalities prevent their simultaneous implementation when applied through the same actuation source, such as the rotor thrusts used by conventional aerial robots. To overcome this limitation, we propose a hybrid impedance--admittance control strategy for a multi-link aerial robot. The articulated morphology enables a functional separation of force and motion regulation across joint and rotor actuation sources. In this framework, admittance behavior is generated through joint angle regulation to enhance adaptive interaction, while impedance behavior is achieved by modulating rotor thrust to regulate the sliding motion. This structural coordination allows the robot to leverage the complementary strengths of both control paradigms. As a result, the multi-link aerial robot achieves resilient and adaptive surface sliding. Experimental results demonstrate robust and compliant sliding performance on unknown surfaces.

\end{abstract}

\section{INTRODUCTION}
\label{sec:introduction}
Multi-link aerial robots are capable of actively deforming their articulated serial-link structures in the air, and this deformability provides additional degrees of freedom for manipulation during flight \cite{zhao_2021_singularity, zhao_2018_design, yang_2018_lasdra, hameed_2025_dragonfly}. Owing to these structural advantages, multi-link aerial robots have great potential for applications such as inspection and search-and-rescue tasks. Their effectiveness has been demonstrated in a variety of experimental tasks, such as grasping objects \cite{ hameed_2025_dragonfly, shi_2020_aerial, zhao_2023_versatile}, opening sliding door \cite{sugito_2022_aerial} and manipulating valves \cite{zhao_2022_forceful}. 
However, before the multi-link aerial robots can be deployed in a broader range of aerial manipulation tasks, they must be able to handle more complex interaction scenarios involving sustained contact and environmental uncertainties.

A representative interaction scenario is a contact-rich surface sliding task, in which a multi-link aerial robot is typically required to maintain a pressing contact between its end-effector and an unknown uneven surface while following a desired sliding trajectory \cite{bodie_2020_active, trujillo_2019_novel, tognon_2019_truly}. 
\begin{figure}[t]
  \centering
  \includegraphics[width=\linewidth]{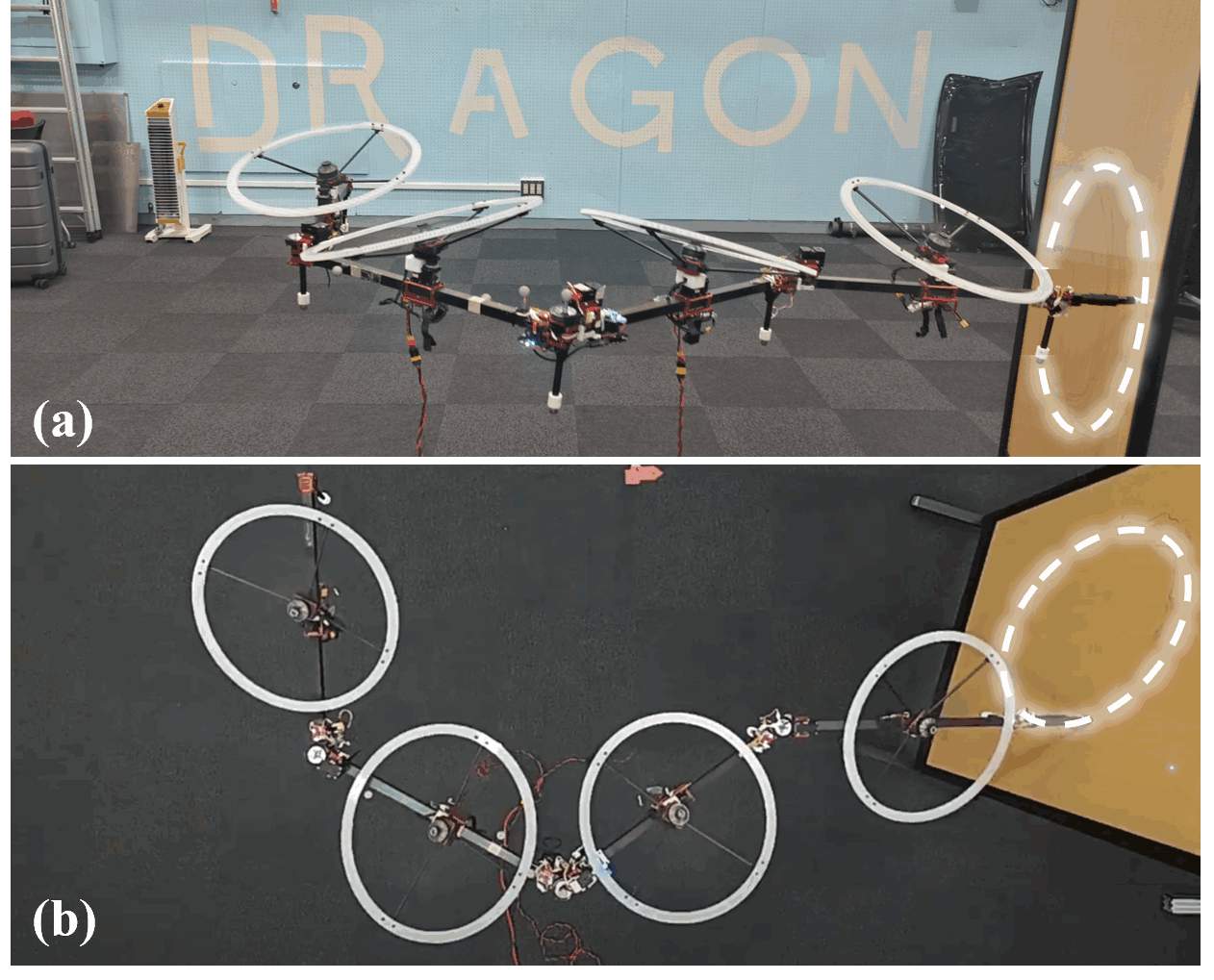}
  \caption{A multi-link aerial robot is tracking a circular trajectory on a sloped unmodeled board while maintaining contact. The desired trajectory is marked by a white dashed line. (a) Side view. (b) Top view.}
  \label{fig:flight}
  \vspace{-7mm}
\end{figure}
Such tasks inherently involve a coupling between unmodeled surface geometry and frictional effects, which leads to significant uncertainties in both the external wrench and the contact geometry experienced by the robot. As a result, the controller must exhibit both robustness and compliant behavior, enabling the robot to remain resilient to force disturbances while adapting to unpredictable changing contact conditions. Nevertheless, existing control methods for multi-link aerial robots remain limited in achieving this unified combination of robustness and compliant interaction. This limitation primarily stems from their lack of perception of external conditions, which prevents the controller from adapting its force and motion responses to uncertainties in the external wrench and contact conditions.

To execute tasks in interaction scenarios, previous work has commonly adopted force-control strategies such as impedance control and admittance control \cite{ollero_2021_past}. Among these strategies, impedance control has been extensively explored for aerial robots. Bodie et al. \cite{bodie_2020_active} introduced a six-DOF axis-selective impedance controller capable of adjusting its impedance gains according to the proximity to the surface, while Zhang et al. \cite{zhang_2022_learning} employed learning-based techniques to adapt impedance parameters according to the interaction conditions. These methods can effectively respond to external force disturbances through gain modulation; however, there is limited evidence that they remain effective in environments with uneven or irregular surfaces. In addition to impedance approaches, admittance control has also been applied in aerial manipulation tasks. Ryll et al. \cite{ryll_2019_6d} enabled an aerial robot to follow variations in the contact-plane geometry, yet the induced lateral force disturbances caused noticeable rotational deviations. Overall, while these studies demonstrate the potential of impedance- or admittance-based approaches for aerial manipulation tasks in interaction scenarios, each method inevitably inherits the limitations of its underlying control principle. 

Impedance control performs well in regulating interaction forces and resisting external force disturbances, but typically exhibits limited adaptability to changing contact geometry. In contrast, admittance control is effective in adapting to geometric variations but generally less robust against large interaction force disturbances. These differences originate from the complementary yet opposite causal relationships of impedance and admittance control paradigms \cite{ott_2010_unified}. Consequently, contact-rich tasks that simultaneously require disturbance rejection and adaptation to uncertain surface geometry cannot be reliably addressed by either approach alone, naturally motivating a combination of both control principles to achieve resilient and adaptive interaction capabilities.
 
However, realizing such a combination is not trivial. Due to the opposite causality of impedance and admittance control, a single actuation source cannot independently realize both interaction objectives and generally requires additional coordination mechanisms \cite{ott_2010_unified}. Conventional aerial robot platforms, which rely exclusively on rotor thrusts, inherently possess only a single actuation source and are therefore limited in their ability to simultaneously exploit the complementary properties of the two paradigms. Although some aerial manipulation platforms (e.g., \cite{lippiello_2012_cartesian, cataldi_2016_impedance}) introduce joint actuation and therefore provide an additional actuation source, existing studies have not leveraged this structural characteristic to simultaneously employ both impedance and admittance control. This limitation further motivates the design of our control strategy.

In this work, we propose a novel hybrid impedance--admittance control enabled by multi-link aerial robots, which exploits the two independent actuation sources of the robot by assigning them distinct control roles. In particular, rotor thrusts govern the global motion of the platform, whereas the joint actuators regulate the end-effector pose relative to the platform’s body. This structure enables impedance control to be implemented through rotor thrusts and admittance control through joint actuation, allowing both paradigms to be realized simultaneously on a single aerial platform. Within the proposed framework, impedance control regulates the aerial robot's center-of-gravity (CoG) motion to achieve disturbance-resilient interaction, while admittance control adjusts joint angles to accommodate unmodeled surface geometric and contact-induced variations. Through this division of roles, the proposed strategy enables resilient and adaptive sliding along unknown surfaces. The task execution process is illustrated in Fig.~\ref{fig:flight}.

The main contributions of this article are:
\begin{enumerate}
\item 
We propose a novel hybrid impedance--admittance controller enabled by a multi-link aerial robot. Impedance behavior is achieved through rotor thrust modulation, whereas admittance behavior at the end-effector is generated by regulating the joint actuators.

\item 
We demonstrate that the proposed control strategy enables resilient and adaptive surface sliding under unmodeled geometric and contact conditions. The strategy maintains disturbance-resilient interaction under friction through impedance control and compensates for surface-induced unmodeled geometric variation via admittance control. 

\item 
We validate the effectiveness of the proposed framework through comparative studies and real-world experiments, showing that it effectively combines the complementary advantages of existing impedance- and admittance-based approaches.
\end{enumerate}

The remainder of the article is organized as follows. The description and modeling for the multi-link aerial robot is introduced in Sec.~\ref{sec:preliminary}. The control strategy is presented in Sec.~\ref{sec:control}. We then show the experimental results in Sec.~\ref{sec:experiments} and finally the conclusion in Sec.~\ref{sec:conclusion}.


\section{Preliminary}
\label{sec:preliminary}

\subsection{Platform Description}

\begin{figure}[t]
    \centering
    \includegraphics[width=\linewidth]{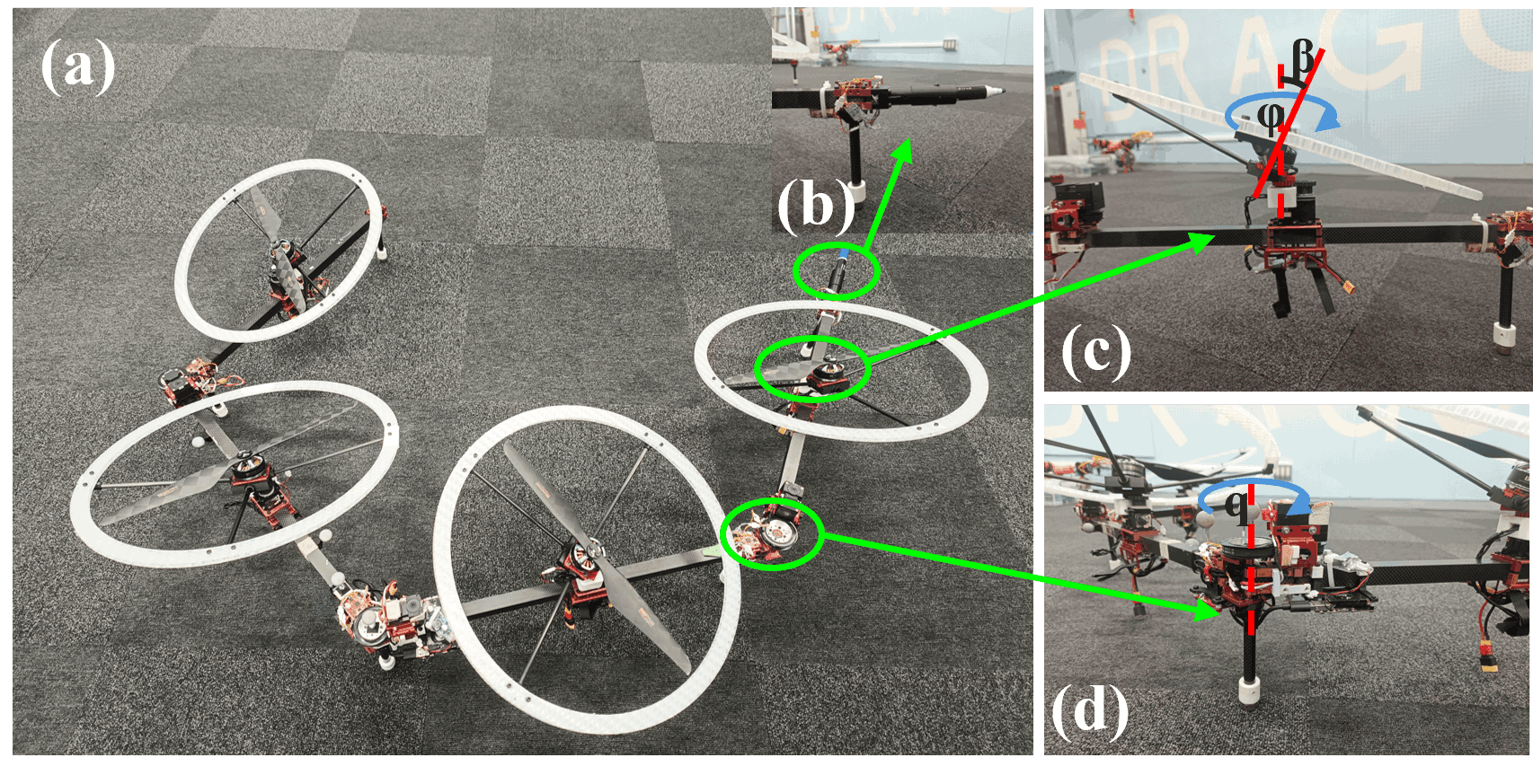}
    \caption{The mechanical design of the multi-link aerial robot. (a) Quad-type multi-link aerial robot used in this work. (b) End-effector. (c) Propeller vectoring module. (d) Joint module.}
    \label{fig:robot}
    \vspace{-5mm}
\end{figure}

The multi-link aerial robot used in this work is shown in Fig.~\ref{fig:robot}. The platform consists of four links, each of which carries an independent propeller unit. The robot employs a serial-link structure with one rotational degree of freedom per joint, whose joint axes are aligned in parallel. Here, $q_i$ is used to represent the $i$th joint angle. Each propeller unit features a fixed tilt angle $\beta$ and a 1-DoF yaw rotation mechanism to vector the thrust direction, whose vectoring angle is denoted by $\psi_i$. The design of the thrust-vectoring unit, shown in the top-right of Fig.~\ref{fig:robot}, was originally developed in \cite{anzai_2019_design}. Additional hardware details can be found in \cite{zhao_2021_singularity}.

\subsection{Modeling}

We denote scalars by regular font $x \in \mathbb{R}$, vectors by bold lowercase $\boldsymbol{x} \in \mathbb{R}^n$, and matrices by bold uppercase $\boldsymbol{X} \in \mathbb{R}^{m \times n}$. A vector expressed in frame $\{A\}$ is written as $^A\boldsymbol{p}$, and the rotation matrix from frame $\{A\}$ to frame $\{B\}$ is denoted as $^A\boldsymbol{R}_B$.
In this paper, $\{W\}$ represents the world frame, $\{C\}$ represents the CoG frame, and $\{F_i\}, i\in\{1,2,3,4\}$ are defined as the coordinate frames of each rotor.

With these definitions, the dynamics of the multi-link aerial robot at the CoG, derived from the Newton--Euler equations, are given by:
\begin{subequations}
\label{eq:dynamics}
\begin{align}
    m(^W\!\ddot{\boldsymbol{p}}+\boldsymbol{g})&= {^W\!\boldsymbol{f}_{cmd}} + {^W\!\boldsymbol{f}_{ext}}, \\[2pt]
    \boldsymbol{I}{^C\!\dot{\boldsymbol{\omega}}}+{^C\!\boldsymbol{\omega}} \times \boldsymbol{I}{^C\!\boldsymbol{\omega}}&={^C\!\boldsymbol{\tau}_{cmd}} + {^C\!\boldsymbol{\tau}_{ext}},
\end{align}
\end{subequations}
where $m$ is the total mass, $\boldsymbol{I}$ is the total inertial matrix expressed in frame $\{C\}$, $^W\boldsymbol{p}$ is the position of the CoG frame $\{C\}$ in the world frame $\{W\}$, $^{C}\boldsymbol{\omega}$ is the angular velocity expressed in the CoG frame $\{C\}$, $\boldsymbol{g}$ is the gravitational acceleration expressed in frame $\{W\}$, ${^W\!\boldsymbol{f}_{cmd}}$, ${^C\!\boldsymbol{\tau}_{cmd}}$ and ${^W\!\boldsymbol{f}_{ext}}$, ${^C\!\boldsymbol{\tau}_{ext}}$ denote the command force and torque, and the external force and torque, respectively. Since the joint and vectoring angles vary during motion, the inertial matrix is configuration dependent and may vary accordingly. Here, we consider the multi-link aerial robot as a quasi-static system, meaning that the variations in the inertial matrix caused by joint and thrust-vectoring motions are negligible within a single control period \cite{zhao_2021_singularity}.

\begin{figure*}[t]
  \centering
  \includegraphics[width=\textwidth]{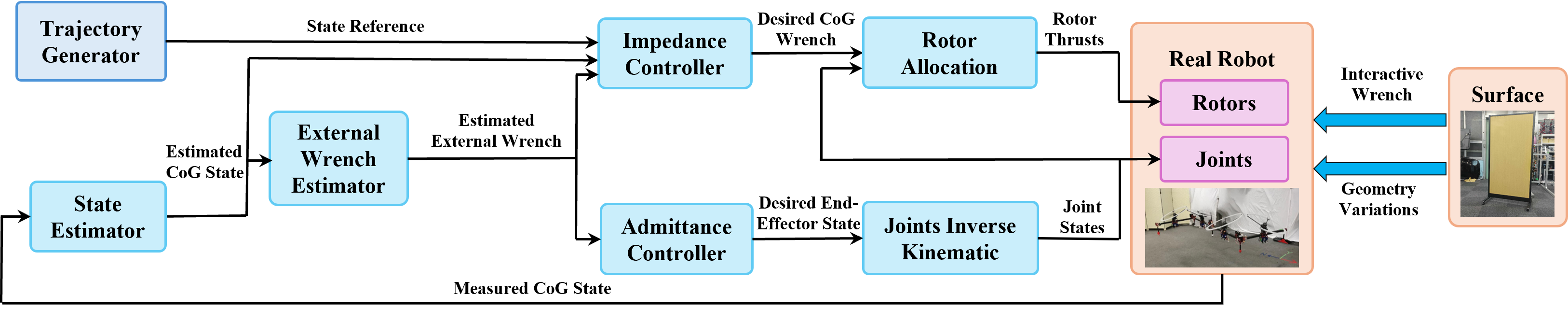}
  \caption{Block diagram of the hybrid impedance--admittance control framework. Inputs for the admittance controller consist of the estimated external wrench. Inputs for the impedance controller include the state reference, estimated CoG state, and estimated external wrench.}
  \label{fig:framework}
   \vspace{-5mm}
\end{figure*}

The command force ${^W\boldsymbol{f}_{cmd}}$ and the command torque ${^{C}\boldsymbol{\tau}_{cmd}}$ are obtained by aggregating the contributions of the four propeller rotors, whose individual forces and torques are expressed in frame $\{C\}$. We compute each rotor's contribution as follows:
\begin{equation}
\label{eq:allocation for force}
    {^C\!\boldsymbol{f}_i}=\lambda_i^C\!\boldsymbol{R}_{F_i}(\boldsymbol{q},\boldsymbol{\psi})\boldsymbol{b}_3,
\end{equation}

\begin{equation}
\begin{aligned}
\label{eq:allocation for torque}
    {^C\!\boldsymbol{\tau}_i}&=^C\!\!\boldsymbol{p}_{F_i}(\boldsymbol{q},\boldsymbol{\psi})\times^C\!\!\boldsymbol{f}_i+\kappa_i ^C\!\boldsymbol{f}_i\\
    &=\lambda_i(^{C}\boldsymbol{\hat{p}}_{F_i}(\boldsymbol{q},\boldsymbol{\psi})+\kappa_i\boldsymbol{E}_3)^C\!\boldsymbol{R}_{F_i}(\boldsymbol{q},\boldsymbol{\psi})\boldsymbol{b}_3,
\end{aligned}
\end{equation}
where $i \in \{1, 2, 3, 4\}$. $\boldsymbol{q} \in \mathbb{R}^3$ and $\boldsymbol{\psi} \in \mathbb{R}^4$ correspond to joint angles and vectoring angles, respectively. The joint angles and vectoring angles determine $^{C}\!\boldsymbol{R}_{F_i}(\boldsymbol{q},\boldsymbol{\psi})\in \mathbb{R}^{3\times3}$ and $^{C}\boldsymbol{p}_{F_i}(\boldsymbol{q},\boldsymbol{\psi})\in \mathbb{R}^3$, which denote the rotation matrix and the position vector of the rotor frame $\{F_i\}$ with respect to the CoG frame $\{C\}$. The rotor frame $\{F_i\}$ is attached to the rotor mount, and the thrust force $\lambda_i$ is along the $z$ axis of $\{F_i\}$. The vector $\boldsymbol{b}_3 = \begin{bmatrix}0 & 0 & 1\end{bmatrix}^{\mathrm{T}}$ is a unit vector and $\kappa_i$ is the coefficient associated with the rotor drag moment. The operator $\hat{\cdot{}}$ converts a vector into a skew-symmetric matrix, and $\boldsymbol{E}_3$ denotes the $3 \times 3$ identity matrix. 

Using \eqref{eq:allocation for force} and \eqref{eq:allocation for torque}, the mapping from the thrust vector $\boldsymbol{\lambda} = \begin{bmatrix}\lambda_1 & \lambda_2 & \lambda_3&\lambda_4\end{bmatrix}^{\mathrm{T}}$ to the command wrench is given by:
\begin{equation}
\label{eq:allocation for wrench}
\begin{pmatrix}
^C\!\boldsymbol{f}_{cmd} \\
^C\!\boldsymbol{\tau}_{cmd}
\end{pmatrix}
=
\sum_{i=1}^{4}
\begin{pmatrix}
^C\!\boldsymbol{f}_i \\
^C\!\boldsymbol{\tau}_i
\end{pmatrix} 
=\begin{pmatrix}
\boldsymbol{Q}_t(\boldsymbol{q},\boldsymbol{\psi}) \\
\boldsymbol{Q}_r(\boldsymbol{q},\boldsymbol{\psi})
\end{pmatrix}\boldsymbol{\lambda},
\end{equation}
where $\boldsymbol{Q}_t(\boldsymbol{q},\boldsymbol{\psi}) \in \mathbb{R}^{3\times4}$ and $\boldsymbol{Q}_r(\boldsymbol{q},\boldsymbol{\psi}) \in \mathbb{R}^{3\times4}$ denote the translational and rotational components of the allocation matrix, respectively. The command force in the world frame can be obtained by $^W\!\boldsymbol{f}_{cmd} = ^W\!\!\!\boldsymbol{R}_C{}^C\!\boldsymbol{f}_{cmd}$.


\section{Control}
\label{sec:control}

\subsection{Framework Overview}

\begin{figure}[h]
    \centering
    \includegraphics[trim={0cm 0.3cm 0cm 0cm}, clip,width=\linewidth]{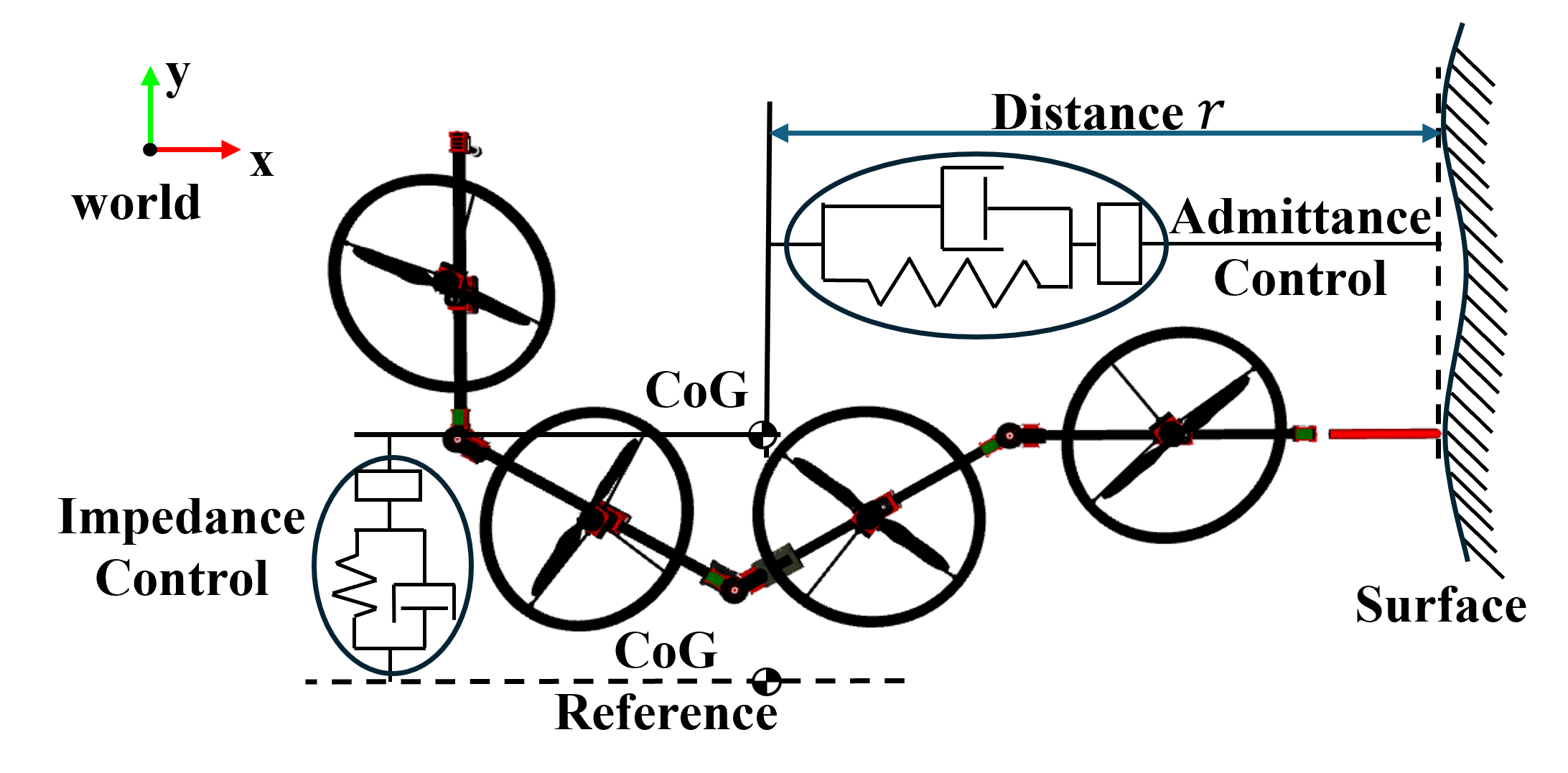}
    \caption{Schematic diagram of the hybrid impedance--admittance control.}
    \label{fig:hybrid_controller}
    \vspace{-5mm}
\end{figure}

The motivation for this hybrid impedance--admittance control framework is to provide both resilience and adaptivity when sliding on an unmodeled surface, where geometry and friction introduce significant uncertainties. As illustrated in Fig.~\ref{fig:framework}, impedance control manages the overall state of the aerial robot, which is represented by the pose of its CoG, while admittance control is applied to regulate the end-effector state, resulting in adjustments of the joint angles. Since variations in joint angles modify the robot’s configuration and thus influence the thrust allocation, this dependency is taken into account in the thrust computation.
As illustrated in Fig.~\ref{fig:hybrid_controller}, by combining both controllers, the multi-link aerial robot can regulate its interaction with the surface in terms of either force or position along different directions. Detailed analysis of the advantages of this hybrid control is presented in the following section.

\subsection{External Wrench Estimation}

Both impedance control and admittance control require an external wrench in their formulations. To estimate the external forces acting on the CoG without introducing additional payload through force sensors, this section presents a momentum-based external wrench estimation method following \cite{ruggiero_2014_impedance}: 

\begin{subequations}
\label{eq:external wrench estimation}
\begin{align}
 {^W\!\boldsymbol{\tilde{f}}_{ext}}=\boldsymbol{K}_{ti}(m^W\!\dot{\boldsymbol{p}}-\int({^W\!\boldsymbol{f}_{cmd}}-m\boldsymbol{g}+ {^W\!\boldsymbol{\tilde{f}}_{ext}})\mathrm{d}t),\\
 {^C\!\boldsymbol{\tilde{\tau}}_{ext}}=\boldsymbol{K}_{ri}(\boldsymbol{I}^C\!\boldsymbol{\omega}-\int({^C\!\boldsymbol{\tau}_{cmd}}-{^C\!\boldsymbol{\omega}} \times \boldsymbol{I}{^C\!\boldsymbol{\omega}} + {^C\!\boldsymbol{\tilde{\tau}}_{ext}})\mathrm{d}t),
\end{align}
\end{subequations}
where ${^W\!\boldsymbol{\tilde{f}}_{ext}}$ and ${^C\!\boldsymbol{\tilde{\tau}}_{ext}}$ denote the estimated external force and torque, respectively, and are assumed to approximate the real external force ${^W\!\boldsymbol{f}}_{ext}$ and the real external torque   ${^C\!\boldsymbol{\tau}_{ext}}$. $\boldsymbol{K}_{ti} \in \mathbb{R}^{3\times3}$ and $\boldsymbol{K}_{ri} \in \mathbb{R}^{3\times3}$are the estimator translational and rotational gain matrices, respectively.

Differentiating (\ref{eq:external wrench estimation}) and substituting (\ref{eq:dynamics}) into it yields a first-order low-pass filtered form: 
\begin{subequations}
\label{eq:external wrench estimation low-pass}
\begin{align}
 {^W\!\boldsymbol{\dot{\tilde{f}}}_{ext}}&=\boldsymbol{K}_{ti}({^W\!\boldsymbol{f}}_{ext}-{^W\!\boldsymbol{\tilde{f}}_{ext}}),\\
 {^C\!\boldsymbol{\dot{\tilde{\tau}}}_{ext}}&=\boldsymbol{K}_{ri}({^C\!\boldsymbol{\tau}_{ext}}-{^C\!\boldsymbol{\tilde{\tau}}_{ext}}).
\end{align}
\end{subequations}

Note that \eqref{eq:external wrench estimation} estimates external forces and torques without using acceleration measurements, relying solely on the estimated linear and angular velocities. By replacing the real external wrench terms in (\ref{eq:dynamics}) with the estimated terms, the controller can be implemented without direct wrench measurements.

\subsection{Hybrid Impedance--Admittance Control}

\subsubsection{\textbf{Impedance Control}}
Impedance control models the robot as a second-order system, whose dynamic parameters can be freely designed to achieve the desired interaction behavior. Therefore, by adjusting the stiffness parameters, the robot can be made resilient in certain directions to resist disturbances, while remaining compliant in others to maintain safe contact interaction \cite{bodie_2020_active}. The translational motion is expressed in the world frame $\{W\}$, whereas the rotational motion is expressed in the CoG frame $\{C\}$. Under this convention, the impedance equations are written as:
\begin{subequations}
\label{eq:impedance control 1}
\begin{align}
    {\boldsymbol{M}_{d}}{^W\!\ddot{\boldsymbol{p}}}+{\boldsymbol{C}_{td}}{\boldsymbol{e}_v}+{\boldsymbol{K}_{td}}{\boldsymbol{e}_p}&={^W\!\boldsymbol{\tilde{f}}_{ext}}, \\[2pt]
    {\boldsymbol{I}_{d}}{^C\!\dot{\boldsymbol{\omega}}}+{\boldsymbol{C}_{rd}}{\boldsymbol{e}_\omega}+{\boldsymbol{K}_{rd}}{\boldsymbol{e}_R}&={^C\!\boldsymbol{\tilde{\tau}}_{ext}},
\end{align}
\end{subequations}
where ${\boldsymbol{M}_{d}}\in \mathbb{R}^{3\times3}$ is the desired virtual mass, ${\boldsymbol{I}_{d}}\in \mathbb{R}^{3\times3}$ is the desired virtual inertia, ${\boldsymbol{C}_{td}}\in \mathbb{R}^{3\times3}$, ${\boldsymbol{C}_{rd}}\in \mathbb{R}^{3\times3}$ are the desired linear and rotational damping matrices, and ${\boldsymbol{K}_{td}}\in \mathbb{R}^{3\times3}$, ${\boldsymbol{K}_{rd}}\in \mathbb{R}^{3\times3}$ are the desired linear and rotational stiffness matrices, respectively. All of these matrices are positive definite. Specifically, the error vectors in (\ref{eq:impedance control 1}) are defined as:
\begin{subequations}
\label{eq:error of rotation}
\begin{align}
    {\boldsymbol{e}_p}&={^W\!\boldsymbol{p}}-{^W\!\boldsymbol{p}_{ref}}, \\[2pt]
    {\boldsymbol{e}_v}&={^W\!\dot{\boldsymbol{p}}}-{^W\!\dot{\boldsymbol{p}}_{ref}}, \\[2pt]
    {\boldsymbol{e}_R}&=\frac{1}{2}({^W\!\boldsymbol{R}^{\mathrm{T}}_{C,ref}}{^W\!\boldsymbol{R}_{C}}-{^W\!\boldsymbol{R}^{\mathrm{T}}_{C}}{^W\!\boldsymbol{R}_{C,ref}})^\vee, \\[2pt]
    {\boldsymbol{e}_\omega}&={^C\!\boldsymbol{\omega}}-{^W\!\boldsymbol{R}^{\mathrm{T}}_{C}}{^W\!\boldsymbol{\omega}}_{ref},
\end{align}
\end{subequations}
where we use the vee-operator $^\vee$ to extract a vector from a skew symmetric matrix.

As we treat the multi-link aerial robot as an underactuated model \cite{zhao_2021_singularity}, the translational and rotational motions are coupled, so the pose control is cascaded. By substituting \eqref{eq:dynamics} into \eqref{eq:impedance control 1} to eliminate $\ddot{\boldsymbol{p}}$ and $\dot{\boldsymbol{\omega}}$, the command wrench added on the CoG can be derived. 

We begin with the translational component:
\begin{equation}
\label{eq:imp cmd for trans1}
\begin{aligned}[t]
    {^W\!\boldsymbol{f}_{cmd}}&=(m\boldsymbol{M}^{-1}_d-\boldsymbol{E}_3){^W\!\boldsymbol{\tilde{f}}_{ext}}\\
    &-m\boldsymbol{M}_d^{-1}({\boldsymbol{C}_{td}}{\boldsymbol{e}_v}+{\boldsymbol{K}_{td}}{\boldsymbol{e}_p})+m\boldsymbol{g}.
\end{aligned}
\end{equation}
To simplify the expression, we define normalized linear damping and stiffness as $\tilde{\boldsymbol{C}}_{td} = \boldsymbol{M}^{-1}_d\boldsymbol{C}_{td}$ and  $\tilde{\boldsymbol{K}}_{td} = \boldsymbol{M}^{-1}_d\boldsymbol{K}_{td}$, respectively. Using these definitions, \eqref{eq:imp cmd for trans1} can be rewritten as
\begin{equation}
\label{eq:imp cmd for trans2}
    {^W\!\boldsymbol{f}_{cmd}}=(m\boldsymbol{M}^{-1}_d-\boldsymbol{E}_3){^W\!\boldsymbol{\tilde{f}}_{ext}}-m(\tilde{\boldsymbol{C}}_{td}{\boldsymbol{e}_v}+{\tilde{\boldsymbol{K}}_{td}}{\boldsymbol{e}_p})+m\boldsymbol{g}.
\end{equation}
To obtain the desired thrust vector, we first calculate the translational command force magnitude by
\begin{equation}
\label{eq:thrust translation 1}
    f_{cmd}=(^W\!\boldsymbol{R}_{C}\boldsymbol{b}_3)\cdot{^W\!\boldsymbol{f}_{cmd}},
\end{equation}
where $^{W}\!\boldsymbol{R}_{C}\boldsymbol{b}_3$ represents the unit vector along the $z$ axis of the CoG frame expressed in the world frame.
Next, we define the following allocation relationship:
\begin{equation}
\label{eq:thrust translation 2}
     \begin{bmatrix} mf_{cmd}&0&0&0\end{bmatrix}^{\mathrm{T}}=\begin{pmatrix}
\boldsymbol{Q}_{tz} \\
\boldsymbol{Q}_r
\end{pmatrix}\boldsymbol{\lambda}_t,
\end{equation}
where $\boldsymbol{Q}_{tz} \in \mathbb{R}^{1 \times 4}$ is the third row of $\boldsymbol{Q}_t$. Solving \eqref{eq:thrust translation 2} yields the translational thrust vector $\boldsymbol{\lambda}_t$. Note that the purpose of the translational command force is to counteract gravity and generate the required translational acceleration, while ensuring that it does not affect the rotational direction.

Next, we determine the thrust vector associated with the rotational component.
The desired roll angle $^W\!\alpha^{tgt}_x$ and desired pitch angle $^W\!\alpha^{tgt}_y$ are computed as
\begin{subequations}
\label{eq:target roll and pitch angle}
\begin{align}
    {^W\!\alpha^{tgt}_x}&=atan^{-1}(-{^W\!\bar{f}_{y,cmd}},\sqrt{{^W\!\bar{f}_{x,cmd}^2}+{^W\!\bar{f}_{z,cmd}^2}}), \\[2pt]
    {^W\!\alpha^{tgt}_y}&=atan^{-1}(^W\!\bar{f}_{x,cmd},{^W\!\bar{f}_{z,cmd}}),
\end{align}
\end{subequations}
where
\begin{equation}
\label{eq:imp cmd for trans}
    \begin{bmatrix}
^W\!\bar{f}_{x,cmd} & ^W\!\bar{f}_{y,cmd} & ^W\!\bar{f}_{z,cmd}
\end{bmatrix}^{\mathrm{T}}=R^{-1}_Z(^W\!\alpha_z)^W\!\boldsymbol{f}_{cmd}.
\end{equation}
Here, $R_Z(\cdot{})$ is a rotation matrix that only rotates along the $z$ axis with a certain angle.
Employing the computed target roll $^W\!\alpha^{tgt}_x$, target pitch $^W\!\alpha^{tgt}_y$, and the reference yaw angle $^W\!\alpha^{ref}_z$, the reference rotation matrix $^W\!R_{C,ref}$ in \eqref{eq:error of rotation} is constructed, from which the rotation error vector ${\boldsymbol{e}_R}$ is derived. 

After getting the error vectors ${\boldsymbol{e}_R}$, we can calculate the command torque
\begin{equation}
\label{eq:imp cmd for rot1}
\begin{aligned}
    {^{C}\boldsymbol{\tau}_{cmd}}&=(\boldsymbol{I}\boldsymbol{I}^{-1}_d-\boldsymbol{E}_3){^{C}}\boldsymbol{\tilde{\tau}}_{ext}-\\
    &\boldsymbol{I}\boldsymbol{I}^{-1}_d({\boldsymbol{C}_{rd}}{\boldsymbol{e}_\omega}+{\boldsymbol{K}_{rd}}{\boldsymbol{e}_R})+{^{C}\boldsymbol{\omega}} \times \boldsymbol{I}{^{C}\boldsymbol{\omega}}.
\end{aligned}
\end{equation}
We define the normalized rotation damping and rotation stiffness as $\tilde{\boldsymbol{C}}_{rd} = \boldsymbol{I}^{-1}_d\boldsymbol{C}_{rd}$ and  $\tilde{\boldsymbol{K}}_{rd} = \boldsymbol{I}^{-1}_d\boldsymbol{K}_{rd}$, respectively. Equation (\ref{eq:imp cmd for rot1}) can then be rewritten as
\begin{equation}
\label{eq:imp cmd for rot2}
\begin{aligned}
    {^{C}\boldsymbol{\tau}_{cmd}}&=(\boldsymbol{I}\boldsymbol{I}^{-1}_d-\boldsymbol{E}_3){^{C}}\boldsymbol{\tilde{\tau}}_{ext}-\\
    &\boldsymbol{I}(\tilde{\boldsymbol{C}}_{rd}{\boldsymbol{e}_\omega}+\tilde{\boldsymbol{K}}_{rd}{\boldsymbol{e}_\alpha})
    +{^{C}\boldsymbol{\omega}} \times \boldsymbol{I}{^{C}\boldsymbol{\omega}}.
\end{aligned}
\end{equation}
Then the rotation thrust vector can be obtained by
\begin{equation}
\label{eq:thrust rotation 1}
     \boldsymbol{\lambda}_r=\boldsymbol{Q}_r^{\dagger}{^{C}\boldsymbol{\tau}_{cmd}},
\end{equation}
where $\boldsymbol{Q}_r^{\dagger}$ is the Moore--Penrose pseudoinverse of $ \boldsymbol{Q}_r$.

Finally, the total thrust required to be generated by the rotors can be obtained as
\begin{equation}
\label{eq:thrust}
\boldsymbol{\lambda}=\boldsymbol{\lambda}_t+\boldsymbol{\lambda}_r.
\end{equation}
This gives the individual thrust values that each rotor needs to generate. The details of the thrust allocation process can be found in \cite{zhao_2021_singularity}.

\subsubsection{\textbf{Admittance Control}}

Similar to impedance control, admittance control also regulates the robot’s response to external forces through a virtual dynamic relationship. However, impedance control generates a resisting force in response to displacement errors, whereas admittance control directly produces a compliant displacement in response to external forces. In this work, we employ admittance control to regulate the end-effector state by adjusting the joint angles in response to the external forces induced by surface geometric variations, thereby mitigating both the force and position disturbances experienced by the CoG through deformation. For the surface sliding task considered in this study, as illustrated in Fig.~\ref{fig:hybrid_controller}, the $x$ axis of the world frame is defined as the contact direction, while the $y$ and $z$ axes represent the sliding directions. Since the geometric variations occur primarily along the contact direction, admittance control is applied only along the x axis. This avoids unnecessary structural motions in non-contact directions that could compromise the stability of the aerial robot. The admittance dynamics along the $x$ axis can then be written as:

\begin{equation}
\label{eq:admittance control}
\begin{aligned}
    m_a(\ddot{r}_{ref}-\ddot{r})+c_a(\dot{r}_{ref}-\dot{r})+k_a(r_{ref}-r)\\
    =\hat{f}_{ext,x}-f_{ref,x},
\end{aligned}
\end{equation}
where $\hat{f}_{ext,x}$ is filtered from the first component of ${^W\!\boldsymbol{\tilde{f}}_{ext}}$, and $f_{ref,x}$ represents the reference external force along the $x$ axis. The position $r$ is defined as the distance between the end effector and the CoG along the $x$ axis, as shown in Fig.~\ref{fig:hybrid_controller}. The reference position $r_{ref}$ corresponds to the distance between the end-effector and the CoG in the $x$ axis at nominal surface-contact configuration, and the reference velocity and acceleration are set as $\dot{r}_{ref}=0$ and $\ddot{r}_{ref}=0$, respectively. Finally, the parameters $m_a$, $c_a$, $k_a$ denote the admittance mass, damping, and stiffness, respectively.

We then employ a discrete-time difference method to compute the distance $r$ at each time step, denoted as $r[k]$. The corresponding acceleration is calculated as: 
\begin{equation}
\label{eq:difference method1}
\ddot{r}[k]=\frac{(\hat{f}_{ext,x}-f_{ref,x})-k_a(r_{ref}-r[k-1])+c_a\dot{r}[k-1]}{m_a}.
\end{equation}
Using this acceleration, the velocity and position can be updated iteratively according to:
\begin{subequations}
\label{eq:difference method2}
\begin{align}
\dot{r}[k]&=\dot{r}[k-1]+\ddot{r}[k]\Delta t, \\[2pt]
r[k]&=r[k-1]+\dot{r}[k]\Delta t,
\end{align}
\end{subequations}
where $\Delta t$ is the control period.


Given the desired end-effector pose, the joint angles are obtained by solving the inverse kinematics (IK). Consequently, admittance behavior is achieved by deforming the multi-link structure in response to external forces along the $x$ axis. As a result, surface geometry variations are reflected as changes in the contact force. The admittance mechanism absorbs part of the disturbances caused by geometric variations and external forces at the joints, significantly reducing those transmitted to the CoG and improving the overall motion behavior of the aerial robot.

\subsection{Hybrid Control Mechanism} \label{sec:analysis}

In surface sliding tasks, the contact and sliding directions exhibit fundamentally different physical characteristics: the sliding directions are dominated by friction-induced disturbances, whereas the contact direction is shaped by geometric variations of the environment. Leveraging this distinction, the hybrid impedance--admittance controller assigns each axis the control behavior best suited to its interaction properties. As illustrated in Fig.~\ref{fig:hybrid_controller}, admittance control is applied along the contact direction ($x$ axis) to regulate geometry-driven variations, while impedance control governs the sliding directions ($y$ and $z$ axes) to accommodate friction-affected motion.


Through this task-aligned division of control responsibilities, the hybrid controller achieves what neither controller can provide alone: robust sliding performance in the sliding directions and adaptive behavior in the contact direction. This directional decomposition is enabled by the actuation structure of the multi-link aerial robot, which allows the two control paradigms to be realized simultaneously through independent actuation sources. This combination is therefore essential for achieving resilient interaction and adaptive contact behavior in surface sliding tasks.


\section{Experiments} 
\label{sec:experiments}

\subsection{Experimental Setup}
\begin{figure}[t]
     \centering
    \includegraphics[trim={0cm 0.5cm 0cm 0cm}, clip, width=\linewidth]{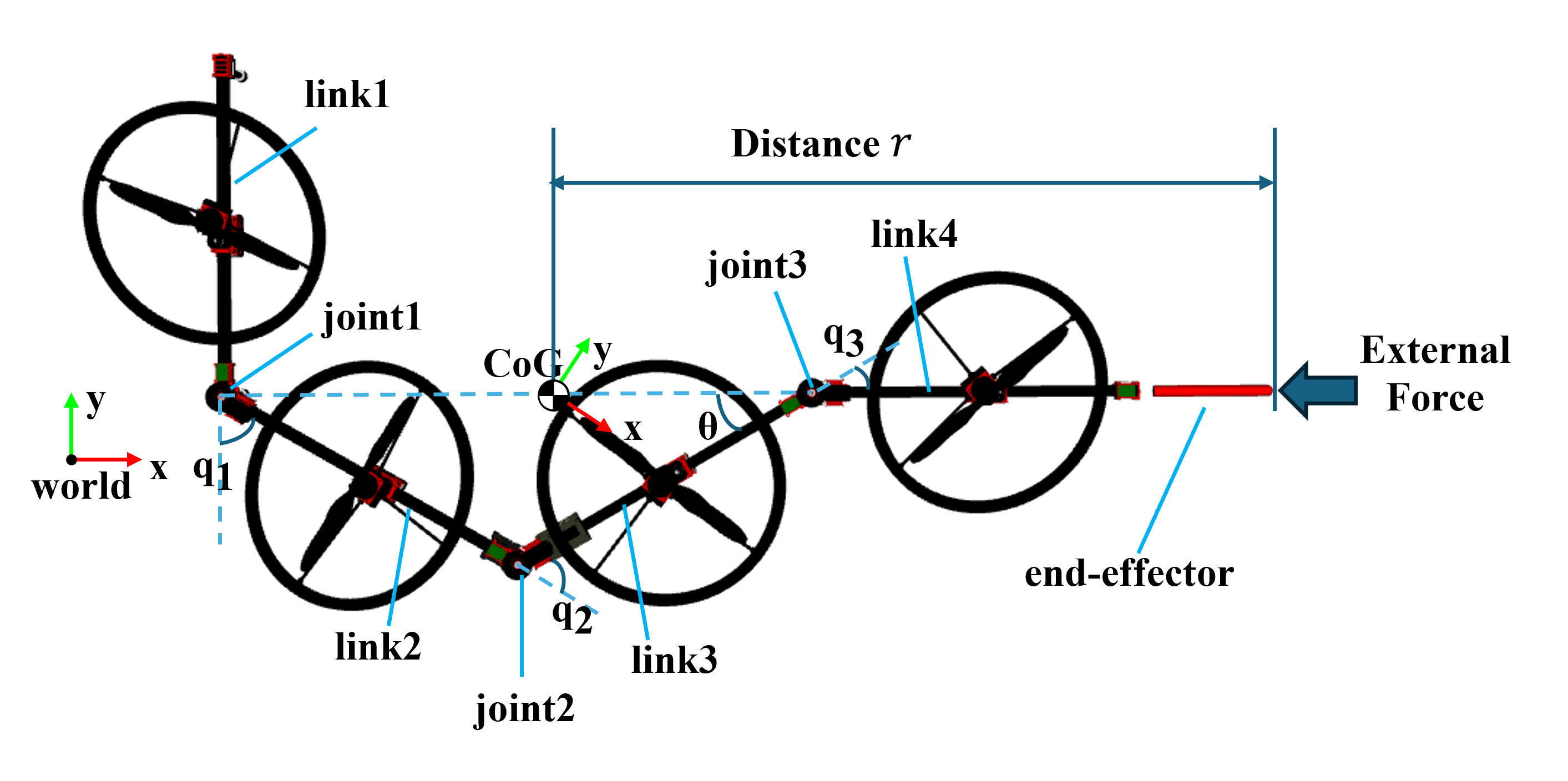}
    \caption{The nominal surface-contact configuration of the multi-link aerial robot. The configuration angle $\theta = \pi /6$, and the joint angles $q_1$, $q_2$, $q_3$ are consequently $\pi /3, \pi /3,-\pi /6$, respectively.}
    \label{fig:admittance control}
    \vspace{-5mm}
\end{figure}
Experiments are conducted using the multi-link aerial robot shown in Fig.~\ref{fig:robot}. The platform is equipped with an UP-Board single-board computer featuring an Intel Atom x5-Z8350 CPU, and the Robot Operating System (ROS) is utilized for communication. A motion capture system provides the full state of the robot throughout the experiments. A marker pen is attached to link 4 as shown in Fig.~\ref{fig:admittance control} serving as the end-effector for contacting with the environment.

The actual experimental parameters are as follows. The estimator translation and rotation gain matrices are given by $\boldsymbol{K}_{ti}=diag(3.0, 3.0, 3.0)$ and $\boldsymbol{K}_{ri}=diag(1.0, 1.0, 1.0)$. We treat $\boldsymbol{M}_d/m$ and $\boldsymbol{I}_d\boldsymbol{I}^{-1}$ as entities. Their values are set to $\boldsymbol{M}_d/m=diag(1.0, 3.0, 1.0)$ and $\boldsymbol{I}_d\boldsymbol{I}^{-1}=diag(10.0, 10.0, 5.0)$. By choosing these values greater than $1$, the impedance controller exhibits more resilient behavior against environmental disturbances due to the increased designed stiffness. The normalized linear stiffness and rotational stiffness are given by $\tilde{\boldsymbol{K}}_{td} = diag(2.5, 2.5, 1.5)$ and $\tilde{\boldsymbol{K}}_{rd} = diag(40.0, 40.0, 10.0)$. The normalized linear and rotational damping terms are not directly specified. Instead, we define the linear damping ratio $\boldsymbol{Z}_{td}=diag(0.6, 0.6, 0.7)$ and the rotational damping ratio $\boldsymbol{Z}_{rd}=diag(1.1, 1.1, 0.8)$, then compute the corresponding normalized damping matrices as follows:
\begin{subequations}
\label{eq:damping rate}
\begin{align}
\tilde{\boldsymbol{C}}_{td}&=2\boldsymbol{Z}_{td}\tilde{\boldsymbol{K}}^{\frac{1}{2}}_{td}, \\[2pt]
\tilde{\boldsymbol{C}}_{rd}&=2\boldsymbol{Z}_{rd}\tilde{\boldsymbol{K}}^{\frac{1}{2}}_{rd}.
\end{align}
\end{subequations}
The admittance mass, damping and stiffness are given by $m_a=40\,\mathrm{kg}$, $c_a=55\,\mathrm{N\cdot s/m}$ and $k_a=16\,\mathrm{N/m}$. The reference external force is given by $f_{ref,x} = 0$.

\subsection{Comparative Study}

As shown in Fig.~\ref{fig:admittance control}, the robot started from its nominal configuration. Throughout the experiment, the yaw angle was adjusted according to the joint configuration such that link 4 remained aligned with the $x$ axis in order to reduce external torque applied on the CoG. Specifically, here we defined the configuration angle $\theta$ as the variable in the admittance controller to determine the configuration solely and then computed the corresponding joint angles analytically through inverse kinematics. The angle of each joint was satisfied as

\begin{equation}
\label{eq:joint angles}
 \begin{bmatrix}
    q_1 \\
    q_2 \\
    q_3
    \end{bmatrix} 
    =
    \begin{bmatrix}
    \frac{\pi}{2}-\theta \\
    2\theta \\
    -\theta
    \end{bmatrix},
\end{equation}
where $\theta$ is shown in Fig.~\ref{fig:admittance control}.

\begin{figure}[t]
    \centering
    \includegraphics[trim={0cm 1cm 0cm 0cm}, clip, width=\linewidth]{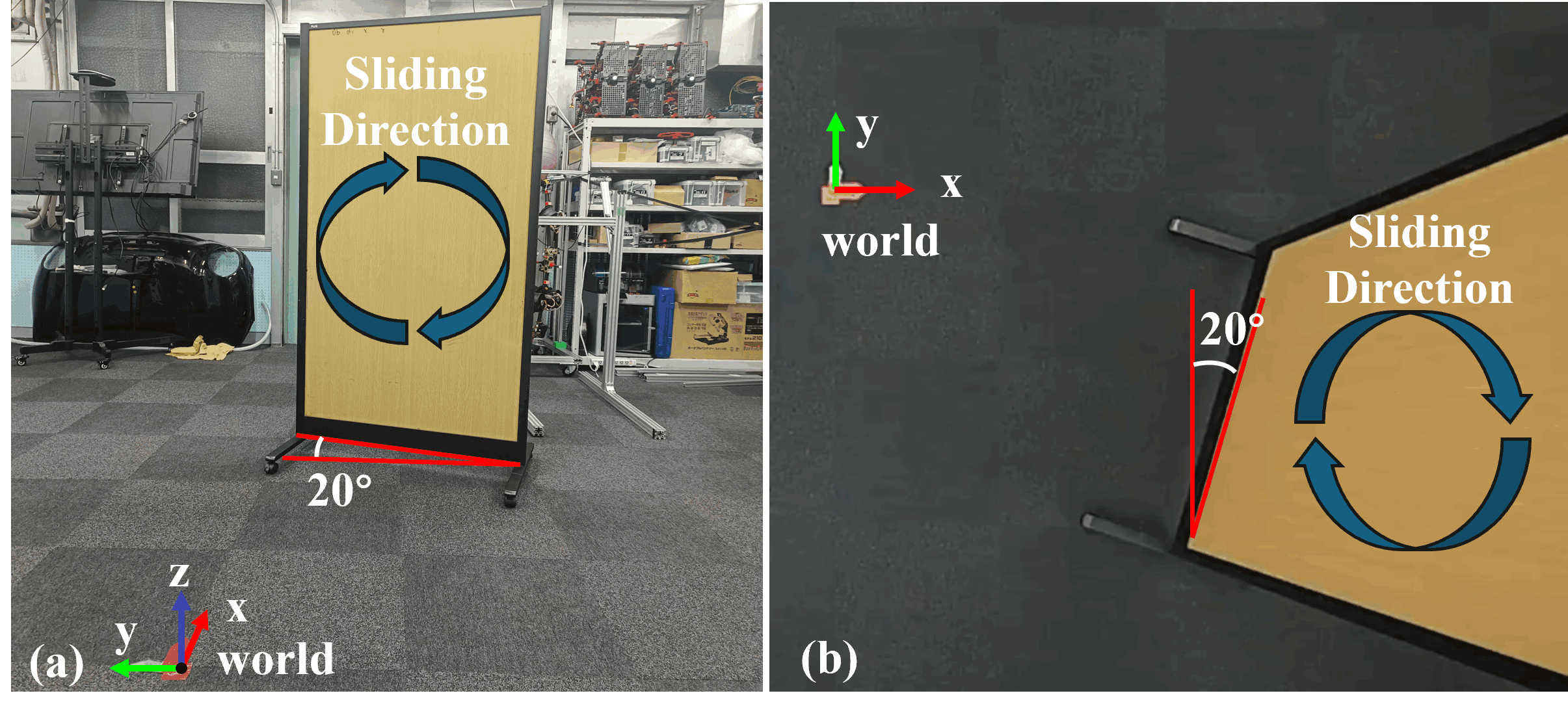}
    \caption{Experiment setup: an board with \textbf{known} position, and \textbf{unknown} sloped angle to the robot controller to mimic an uncertain surface geometric variation. The world frame is marked on the figure. (A) Front view. (b) Top view.}
    \label{fig:exp_scene}
    \vspace{-5mm}
\end{figure}

The experimental environment is illustrated in Fig.~\ref{fig:exp_scene}. A vertical board was placed at a known location within the testing area. The board’s surface provided sufficient friction for stable contact while still allowing smooth sliding of the end-effector (marker pen). The board’s surface was intentionally sloped by approximately $20^\circ$ with respect to the $y$ axis, and the inclination angle was kept unknown to the robot. Although the surface is planar, its unknown inclination introduces unmodeled geometric variations that serve the same role as curved surfaces in this study. The friction coefficient is also unknown. Consequently, both the contact forces and the surface geometry remain unmodeled in this experiment, which provides a challenging benchmark for evaluating the robot's adaptive sliding capability.

During the experiment, the reference trajectory of the multi-link aerial robot was defined on a plane perpendicular to the $x$ axis. The robot was commanded to follow the predefined circular trajectory with a radius of $0.30\,\mathrm{m}$ and a period of $14\,\mathrm{s}$ within this plane. Due to the board’s unknown inclination angle, the desired circular trajectory experienced approximately $0.15\,\mathrm{m}$ of variation along the $x$ axis during one cycle, introducing measurable geometry-induced deviations during contact. This geometric variation was not incorporated into the reference trajectory generation or controller design. Therefore, the robot was required to adapt to this geometric variation online during contact.

To evaluate performance,  comparative experiments were conducted between the proposed hybrid controller and a manually tuned PID baseline. The PID gains were selected for the best stable contact tracking performance. In the hybrid controller, impedance control was applied along the $y$ and $z$ axes to improve sliding resilience, while admittance control was applied along the $x$ axis to enable adaptive surface interaction, with each controller operating independently. To demonstrate the necessity of of the hybrid design, we additionally performed experiments using only impedance control (without admittance-based joint adaptation) and only admittance control (where the CoG state was regulated solely by PID).

\begin{figure*}[t]
    \centering
    \includegraphics[trim={1.5cm 2cm 0cm 0cm}, clip, width=\textwidth]{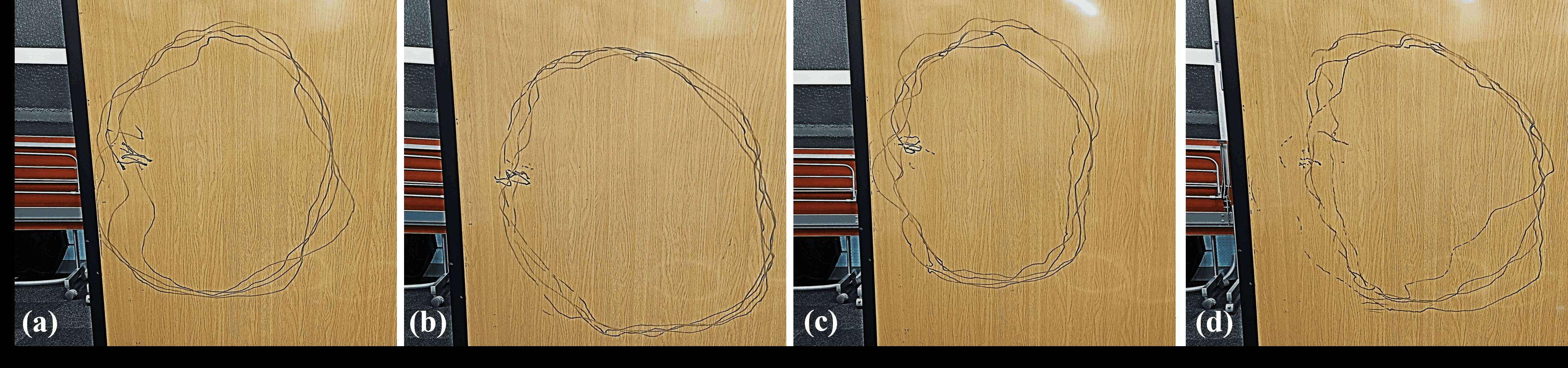}
    \caption{Tracking trajectory on the board. (a) PID controller. (b) Only impedance controller. (c) Only admittance controller. (d) Hybrid impedance--admittance controller.}
    \label{fig:impedance_result}
     \vspace*{-3mm}
\end{figure*}
\begin{figure}[t]
    \centerline{\includegraphics[trim=0 0 0 0,clip,width=3.4in]
    {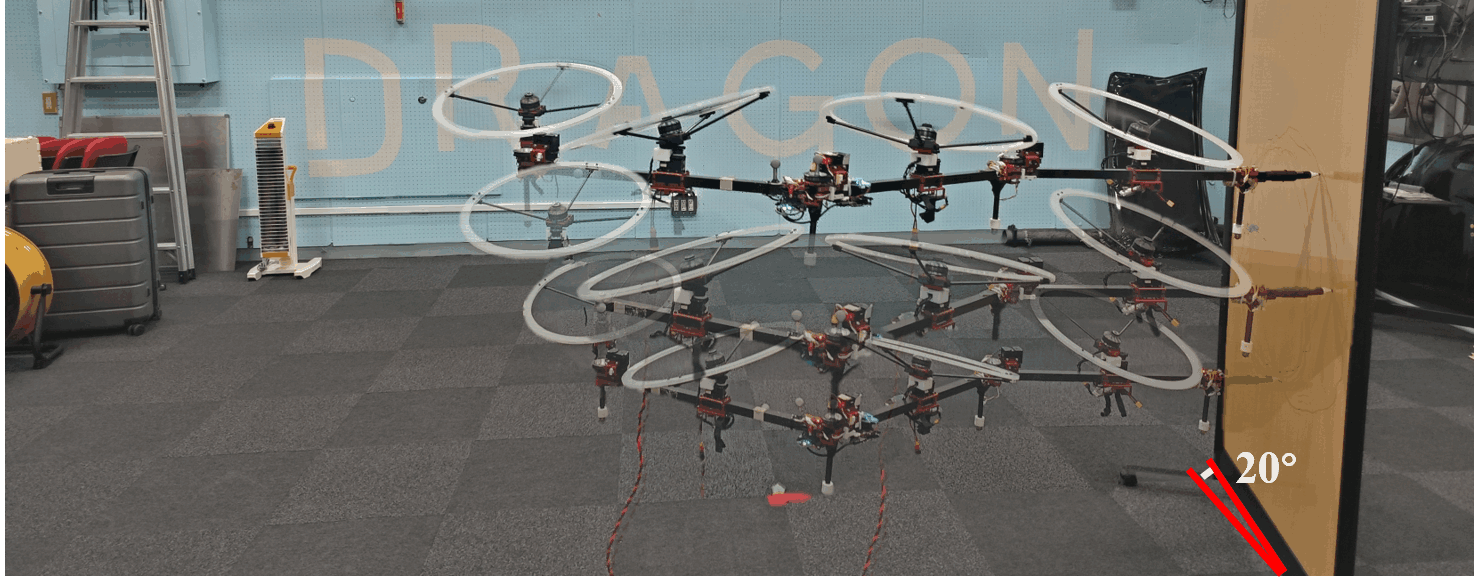}}
    \vspace*{-3mm}
    \caption{The hybrid controller enabled the multi-link aerial robot to deform appropriately when encountering surface geometric variation.}
    \label{fig:admittance_exp}
    \vspace*{-5mm}
\end{figure}

Following this setup, four sets of experiments were performed. Fig.~\ref{fig:impedance_result} shows the final trajectories drawn on the board. Overall, the only impedance controller produced a trajectory closest to the desired circle, while the hybrid controller showed a nearly circular path. In contrast, only admittance controller and PID controller resulted in noticeably smaller and more distorted trajectories. Fig.~\ref{fig:admittance_exp} further illustrates how the multi-link aerial robot deformed to accommodate geometric variations during contact under hybrid controller.

\subsubsection{\textbf{Compare on Impedance-Related Directions}}

\begin{figure}[t]
    \centerline{\includegraphics[trim=0 0 0 0,clip,width=3.4in]
    {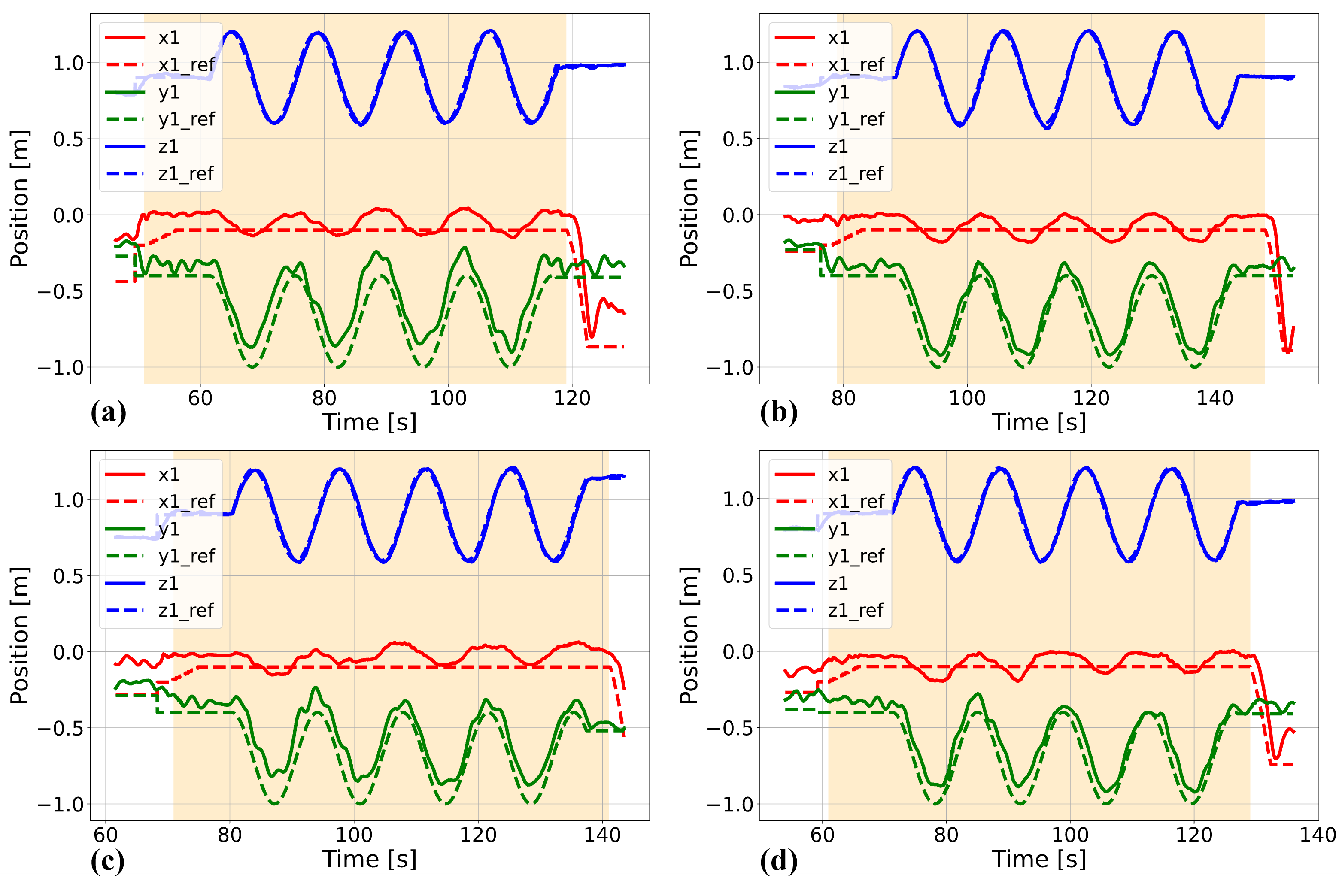}}
    \vspace*{-3mm}
    \caption{Circle trajectory tracking result of the CoG under four controllers. (a) PID controller. (b) Only impedance controller. (c) Only admittance controller. (d) Hybrid impedance--admittance controller. The yellow region denotes the contacting period. }
    \label{fig:impedance_plot}
    \vspace{-4mm}
\end{figure}

\begin{figure}[t]
    \centerline{\includegraphics[trim=0 0 0 0,clip,width=3.4in]
    {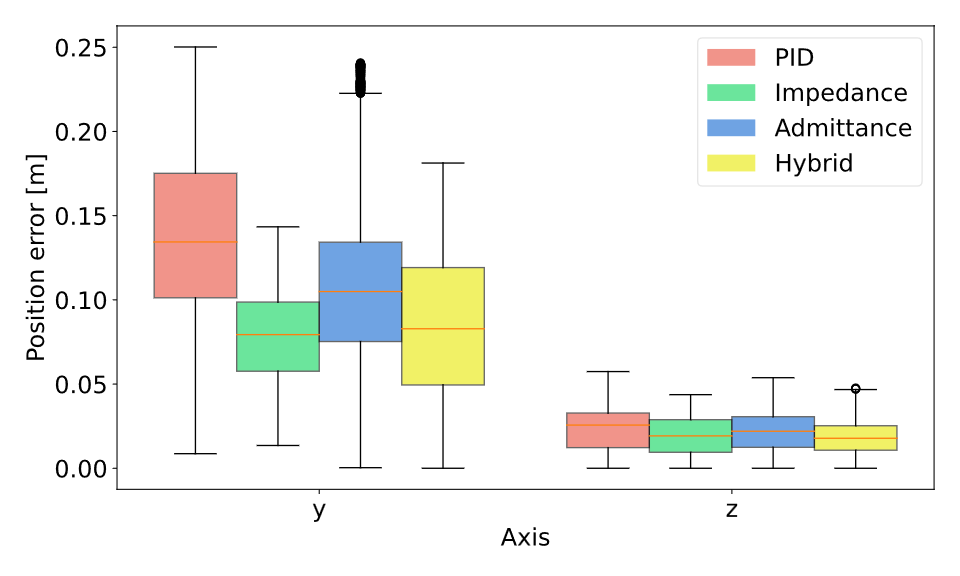}}
    \vspace*{-3mm}
    \caption{Box plots of position errors along the sliding directions (the $y$ and $z$ axes) under the four controllers during the contact period.}
    \label{fig:impedance_box}
    \vspace{-7mm}
\end{figure}

Fig.~\ref{fig:impedance_plot} presents the CoG trajectories together with the reference path under the four controllers. Note that during the sliding task, the $y$ and $z$ axes corresponded to the sliding directions, where friction-induced disturbances were dominant. Although Fig.~\ref{fig:impedance_plot} illustrates the overall trajectories of the four controllers, their performance differences become more evident in the tracking error distributions shown in Fig.~\ref{fig:impedance_box}. The figure shows the distributions of tracking errors of the CoG in the $y$ and $z$ axes. In particular, the median $y$ axis errors were approximately $0.134\,\mathrm{m}$ for PID controller, $0.079\,\mathrm{m}$ for only impedance controller, $0.105\,\mathrm{m}$ for only admittance controller, and $0.083\,\mathrm{m}$ for the hybrid controller. These results indicate that impedance-based methods provided more stable tracking performance than the conventional PID controller when sliding under friction.

This part of the experiments shows that along the sliding directions, impedance controller provided a resilient response to frictional disturbances, enabling more accurate sliding without requiring explicit friction values while still approximately following the desired trajectory.

\subsubsection{\textbf{Compare on Admittance-Related Direction}}

\begin{figure}[t]
    \centerline{\includegraphics[trim=0 0 0 0,clip,width=3.4in]{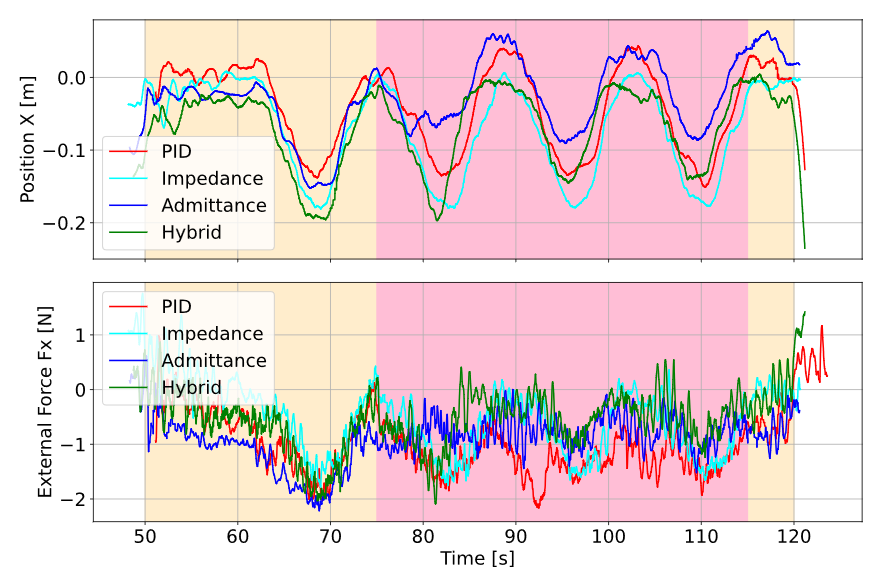}}
    \vspace*{-3mm}
    \caption{Plot of position variation and external force on the CoG in the $x$ axis. The yellow region denotes the contact period, and the pink region indicates the interval during which the admittance controller actively adjusts the configuration.}
    \label{fig:admittance_plot}
    \vspace*{-3mm}
\end{figure}

\begin{figure}[t]
    \centerline{\includegraphics[trim=0 0 0 0,clip,width=3.4in]
    {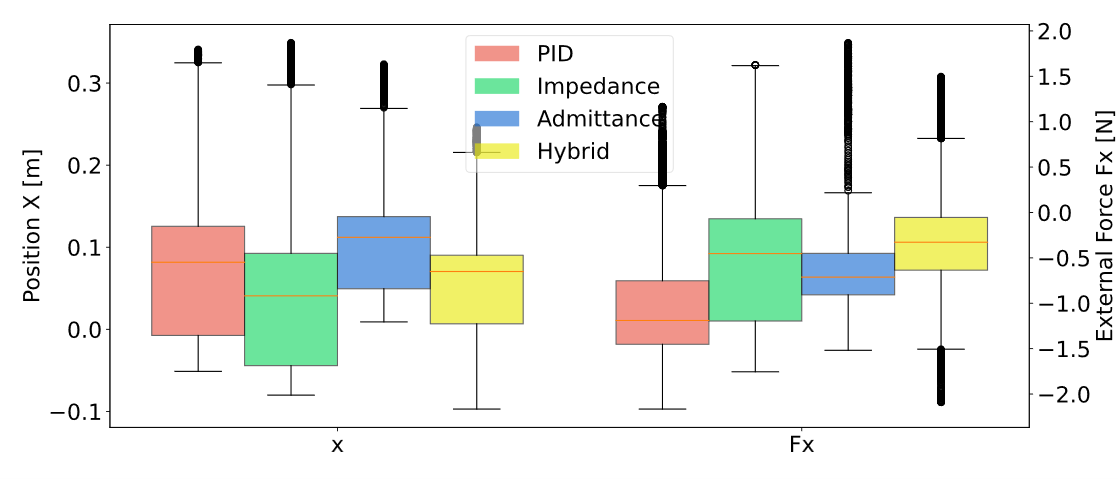}}
    \vspace*{-3mm}
    \caption{Box plots of CoG position and CoG external force along the contact direction ($x$ axis) for the four controllers during the period when the admittance controller was active.}
    \label{fig:admittance_box}
    \vspace*{-5mm}
\end{figure}

Fig.~\ref{fig:admittance_plot} shows the CoG displacement and external force along the $x$ axis. As further quantified in Fig.~\ref{fig:admittance_box}, the PID and only impedance controllers produced similar CoG displacements ($0.133\,\mathrm{m}$ and $0.137\,\mathrm{m}$), both largely reflecting the board’s deformation. In contrast, the admittance and hybrid controllers reduced the CoG displacement to $0.088\,\mathrm{m}$ and $0.084\,\mathrm{m}$, demonstrating their effectiveness in mitigating geometric variation effects.

Fig.~\ref{fig:admittance_box} also shows that the median external forces on the CoG were approximately $-1.190\,\mathrm{N}$ for the PID controller, $-0.453\,\mathrm{N}$ for the only impedance controller, $-0.712\,\mathrm{N}$ for the only admittance controller, and $-0.328\,\mathrm{N}$ for the hybrid controller. The hybrid controller achieved the smallest median external force. In addition, the admittance and hybrid controllers exhibited relatively narrower force distributions than the PID controller, indicating more stable and adaptive interactions with the uncertain surface.

These improvements demonstrate that the admittance controller enables the robot to adjust its configuration in response to surface geometric variations, thereby reducing the influence of geometric variations on the CoG and improving the contact force behavior.




\subsection{Discussion}

The experiment highlights the complementary roles of the two strategies: impedance control improves sliding smoothness and reduces tracking error, whereas admittance control enhances adaptive interaction by decreasing CoG displacement and improving contact behavior. By combining these two mechanisms, the hybrid controller achieves smaller trajectory deviations, reduced unintended CoG drift, and lower interaction forces simultaneously compared with the baseline PID controller, confirming its effectiveness for compliant contact-rich tasks.

\section{Conclusion} \label{sec:conclusion}

This paper proposed a hybrid impedance--admittance control framework for a multi-link aerial robot performing surface sliding tasks. By assigning impedance control to the force-sensitive sliding directions and admittance control to the geometry-sensitive contact direction, the method exploits the distinct interaction characteristics of each axis. Experiments demonstrated improved trajectory tracking, reduced contact forces, and smaller geometry-induced variation compared with a standard PID controller, validating its effectiveness for surface interaction.

In future work, we aim to extend the framework to an overactuated multi-link aerial robot, which introduces higher-dimensional admittance behavior and requires handling spatial contact forces. We also plan to validate the method on more complex curved surfaces with larger geometric variations and spatially varying friction.









\bibliographystyle{IEEEtran}
\bibliography{bibtex/reference}
\end{document}